# VISION AND LANGUAGE: NOVEL REPRESENTATIONS AND ARTIFICIAL INTELLIGENCE FOR DRIVING SCENE SAFETY ASSESSMENT AND AUTONOMOUS VEHICLE PLANNING

Ross Greer[1, 2], Maitrayee Keskar[1, 2], Angel Martinez-Sanchez[1], Parthib Roy[1], Shashank Shriram[1],
Mohan Trivedi[2]
[1] Machine Intelligence, Interaction, and Imagination (Mi³) Laboratory, University of California, Merced, USA
[2] Laboratory for Intelligent and Safe Automobiles (LISA), University of California, San Diego, USA

**ABSTRACT**
Vision–language models (VLMs) have recently emerged as powerful representation learning systems that align visual observations with natural language concepts, offering new opportunities for semantic reasoning in safety-critical autonomous driving systems. This paper investigates how vision–language representations can support driving scene safety assessment and autonomous vehicle decision-making when carefully integrated into perception, prediction, and planning pipelines. We study three complementary system-level use cases. First, we introduce a lightweight, category-agnostic hazard screening approach that leverages CLIP-based image–text similarity to produce a low-latency semantic hazard signal, enabling robust detection of diverse and out-of-distribution road hazards without explicit object detection or visual question answering. Second, we examine the integration of scene-level vision–language embeddings into a transformer-based trajectory planning framework using the Waymo End-to-End Driving Dataset, analyzing how global semantic representations interact with geometry-aware motion planning. Our results show that naïvely conditioning planners on global vision–language embeddings does not improve trajectory accuracy, highlighting the importance of representation–task alignment and structured grounding for safety-critical planning and motivating development of methods for task-informed extraction of relevant semantic content from vision-language embeddings. Third, we investigate natural language as an explicit, human-authored behavioral constraint on motion planning using the doScenes dataset and an instruction-conditioned planning framework. In this setting, passenger-style instructions grounded in visual scene elements suppress rare but severe planning failures and improve safety-aligned behavior in ambiguous scenarios. Taken together, these findings demonstrate that vision–language representations hold significant promise for autonomous driving safety when used to express semantic risk, intent, and behavioral constraints, and that realizing this potential is fundamentally an engineering problem requiring careful system design rather than direct feature injection.

**INTRODUCTION**
Autonomous driving systems have achieved strong performance in structured, well-defined environments, yet ensuring safety in open-world conditions remains a fundamental challenge. Real-world driving is governed not only by geometry, kinematics, and traffic rules, but by rich semantic context that humans interpret continuously: a temporary construction zone marked by improvised signage, a pedestrian hesitating at a curb, an emergency vehicle stopped in an unusual location, or a passenger saying "stop here by the person" in a situation that admits multiple plausible actions. Many of the most consequential safety decisions in driving arise not from the detection of a specific object, but from how contextual information is interpreted and acted upon.

Traditional autonomous driving perception pipelines have been designed primarily around closed-world assumptions. Perception systems are trained to recognize predefined object classes, prediction models forecast motion based on observed trajectories and map structure, and planners generate continuous trajectories that optimize cost functions under explicit constraints. These approaches perform well in nominal conditions, but they often struggle in situations involving ambiguity, novelty, or human intent—precisely the conditions that dominate safety-critical edge cases. As autonomous vehicles are increasingly deployed in open-world environments, the ability to reason about semantic risk, intent, and context becomes as important as geometric precision.

Recent advances in foundation models, particularly vision–language models (VLMs), introduce a new representational layer for autonomous systems. Models such as CLIP [9] and large multimodal transformers learn joint embeddings that align visual observations with natural language concepts, enabling associations between scenes and abstract notions such as "hazard," "blocked roadway," "yielding situation," or "unsafe to proceed," even when these concepts manifest in visually diverse or previously unseen forms [17]. These representations offer an appealing property for safety-critical systems: they encode semantic meaning in a way that is naturally aligned with how humans describe, reason about, and communicate risk.

From a safety perspective, vision–language representations suggest a shift in how autonomous vehicles might reason about their environment. Rather than relying exclusively on object-centric detections or hand-engineered rules, systems could leverage semantic representations that generalize across scenarios, support human interpretability, and provide a common interface between perception, prediction, planning, and human interaction. However, the introduction of such representations also raises an important engineering question: how should vision–language representations be integrated into autonomous driving systems to improve safety, rather than inadvertently introduce misalignment or ambiguity?

Critically, vision–language representations differ fundamentally from the geometric and kinematic features traditionally used in motion planning. They are global, abstract, and semantic, rather than spatially localized or explicitly relational. As a result, their utility is highly sensitive to where they are introduced in the autonomy stack and how their outputs are interpreted. Treating vision–language embeddings as generic features risks conflating semantic understanding with interpretable control, while ignoring the need for grounding, structure, and task alignment. Using these models safely therefore requires careful engineering design rather than naïve integration.

Most prior work incorporating language into autonomous driving has focused on semantic understanding tasks, such as visual question answering, scene description, or object referral [7, 18-22]. These benchmarks establish that models can associate language with visual content and reason about traffic scenes at a descriptive level. However, they stop short of evaluating whether vision–language representations can change vehicle behavior in safety-relevant ways—such as braking, yielding, waiting, or selecting a safer trajectory in ambiguous situations. As a result, there remains limited empirical understanding of how language-aligned representations influence perception, prediction, and planning in real driving pipelines.

In this paper, we investigate the role of vision–language representations in autonomous driving safety through three complementary system-level case studies that span the perception–planning continuum. First, we study vision–language models as lightweight, category-agnostic hazard indicators, using natural language descriptors to produce a low-latency semantic signal that reflects scene-level risk. This approach treats hazard awareness as a property of the entire scene, enabling generalization to unexpected or out-of-distribution hazards without requiring explicit object detection or localization. Second, we examine the integration of scene-level vision–language embeddings into a trajectory prediction framework, exploring how high-level semantic representations interact with geometry-aware planning models in complex driving scenarios. Third, we investigate natural language as an explicit, human-authored constraint on motion planning, where passenger-style instructions grounded in visual scene elements guide short-horizon vehicle behavior.

These three systems are not intended as competing solutions, but as probes into different roles that vision–language representations can play in safety-critical autonomous driving. Together, they highlight a unifying design principle: vision–language representations are most effective when used to express semantic risk, intent, and behavioral constraints, rather than as undifferentiated features for low-level trajectory generation. When aligned with their natural strengths—semantic abstraction, human interpretability, and flexibility of expression—these representations can meaningfully shape vehicle behavior, particularly in ambiguous or underspecified situations that challenge purely vision-based systems.

Our results underscore that realizing the safety benefits of vision–language representations is an engineering problem, not a modeling shortcut. Careful consideration must be given to representation scope, grounding, and interaction with geometry-aware modules. When appropriately integrated, vision–language representations can suppress unsafe behaviors, improve robustness to rare events, and provide a natural interface for human–vehicle interaction. More broadly, this work argues that safe autonomous vehicles will require representations that bridge human-interpretable semantics and machine-executable planning. As autonomous systems move toward open-world deployment, the ability to reason about intent, context, and risk—using representations aligned with human communication—will be essential. Vision–language models offer a powerful foundation for this capability, provided they are integrated with the constraints and validation demanded by safety-critical systems.

**RELATED RESEARCH**
**Vision-Language Understanding and Grounding for Driving**
Recent work has introduced driving-focused vision–language benchmarks to test whether models can interpret traffic scenes in safety-relevant ways, e.g., answering questions about right-of-way, hazards, and agent intent, often with justification or multi-step reasoning. A representative trend is to evaluate VLMs through driving VQA, where questions probe not only object recognition but also contextual reasoning and "what should happen next." LingoQA [7] operationalizes this direction with large-scale video-based driving QA and emphasizes the gap between model and human performance, motivating better evaluation metrics aligned with human judgment. A closely related near-neighbor is DriveLM [13], which extends single-turn VQA toward graph-structured reasoning that more explicitly connects perception, interaction understanding, and planning-relevant questions.

Importantly, while these benchmarks demonstrate that models can map between language and visual scene content, they typically evaluate descriptive correctness or reasoning consistency rather than whether representations reliably translate into behavioral improvements (e.g., safer braking, yielding, or trajectory choice). This gap motivates our system-level focus on (i) lightweight semantic hazard signals and (ii) language as an explicit behavioral constraint in planning.

Language grounding / referring expressions in driving has also been studied through passenger-to-vehicle command datasets, where instructions refer to entities in the scene and the task is to localize the referred object or region. Talk2Car[2] is a canonical example, framing free-form passenger requests grounded in urban driving imagery. However, object referral alone does not establish that grounded language changes vehicle motion safely; our emphasis is on downstream planning behavior under uncertainty and ambiguity.

**Language-Guided Decision-Making**
A growing body of work explores language as an interface for driving decisions, typically in two regimes [26-28]. First, language as a high-level reasoning layer uses LLM/VLM outputs (e.g., "yield," "stop," "merge") that are executed by downstream planners, improving interpretability but leaving a translation gap between abstract language and continuous control [26, 27]. DiLu [3] exemplifies this paradigm using LLM-centered reasoning and reflection for decision-making in driving environments.

A related direction reduces the abstraction gap between language and motion by translating natural language into more structured, program-like decision processes that can be validated or executed over driving state representations. LaMPilot [25] introduces an open benchmark for autonomous driving with language, framing language as an interface for specifying driving intent and constraints while enabling systematic evaluation of language-guided driving behaviors. By treating language guidance as an executable or verifiable component rather than free-form reasoning alone, this line of work highlights a practical pathway for improving reliability and interpretability of language-in-the-loop autonomy.

A closely related concern is the potential misalignment between language-based reasoning and actual driving decisions. Recent work has shown that even when a language model produces plausible or logically consistent reasoning, the resulting action or trajectory may still be unsafe or suboptimal. RDA-Driver [29] explicitly studies this failure mode by analyzing discrepancies between chain-of-thought reasoning and downstream driving decisions, and proposes reasoning–decision alignment mechanisms to improve planning reliability. By demonstrating that correct or fluent language reasoning does not guarantee correct driving behavior, this work highlights a fundamental limitation of treating language outputs as directly actionable decisions, and reinforces the need for structured grounding and safety arbitration when integrating language into autonomous driving systems.

Second, language as direct conditioning in closed-loop driving injects linguistic inputs into the policy/planning model so language can influence trajectories over time. LMDrive [5] is a representative peer-reviewed example, integrating language with multimodal sensing in a closed-loop framework.

In contrast to both regimes, we evaluate language in a more "systems" framing: (i) as a lightweight semantic hazard score that can act as a screening layer, and (ii) as a short-horizon behavioral constraint that can suppress rare but severe planning failures, while explicitly discussing trust boundaries and arbitration.

**Vision-Language as High-Level Reasoning for Decision and Intent Generation**
A common approach in language-guided autonomous driving uses large language or vision–language models as high-level reasoning modules that output symbolic driving decisions or intents, which are then executed by conventional planning and control systems. In this paradigm, language models process visual and contextual inputs to produce abstract actions such as stop, yield, or change lanes, or intermediate reasoning traces that inform downstream planners.

DiLu [3] exemplifies this approach by treating the language model as a central reasoning component that interprets the driving scene and generates high-level decisions, which are subsequently translated into motion using traditional planning and control pipelines. Similarly, OpenEMMA [4] employs multimodal prompting to guide driving behavior via language-centric reasoning, with downstream modules responsible for converting these outputs into executable trajectories.

Using language for high-level decision-making offers improved interpretability, flexible instruction following, and access to commonsense reasoning that is difficult to encode in hand-designed cost functions. However, these systems face a fundamental limitation: the translation of abstract linguistic decisions into precise, safety-critical motion constraints is often underspecified. As a result, safety and reliability depend primarily on the downstream planner, and errors or ambiguities in language-level reasoning can propagate unpredictably into vehicle behavior.

**Language as Direct Conditioning for Closed-Loop Driving**
A smaller but growing line of work integrates language directly into closed-loop driving models, allowing linguistic inputs to influence the generated trajectories or control actions without an explicit intermediate decision layer. In these systems, language is not used solely for high-level reasoning or intent inference, but instead conditions the driving policy itself, directly affecting vehicle behavior over time.

LMDrive [5] and SimLingo [24] exemplifies this approach by incorporating language prompts into an end-to-end closed-loop driving framework, enabling the model to adjust its driving behavior in response to natural language guidance. By conditioning trajectory generation directly on linguistic inputs, such methods reduce the abstraction gap between language and motion compared to symbolic decision-based pipelines.

While this formulation more directly couples language with vehicle behavior, it also introduces new challenges related to reliability, grounding, and safety. Because language influences low-level actions directly, errors in interpretation or grounding can have immediate behavioral consequences, highlighting the need for careful design and evaluation in safety-critical settings.

**Vision-Language for Safety Assessment and Hazard Awareness**
A parallel line of research addresses the challenge of identifying hazards that fall outside predefined object categories. Most autonomous driving datasets emphasize frequent object classes and structured scenarios, which limits their ability to probe system behavior under rare or unexpected conditions. Novelty-oriented benchmarks such as Lost and Found [10], and SegmentMeIfYouCan [11] expose this gap by focusing on uncommon, ambiguous, or visually subtle obstacles that are poorly represented in standard training distributions.

The COOOL [8] (Challenge Of Out-Of-Label) benchmark extends this direction by explicitly framing hazard perception as an open-vocabulary problem in real-world dashcam video. Rather than assuming a fixed label set, COOOL emphasizes hazards that are rare, low-resolution, or semantically ambiguous, including objects such as animals, debris, smoke, or distant obstacles that may occupy only a small fraction of the image. Many hazards appear below 50×50 pixels or are visually indistinct at the moment a driver begins to react, highlighting the difficulty of early and reliable hazard recognition under realistic conditions.

COOOL emphasizes situations where autonomous systems are forced to reason under uncertainty rather than rely on familiar object categories. This framing motivates the use of language not merely as a descriptive interface, but as a mechanism for expressing uncertainty, assessing risk, and supporting safety-critical decisions in open-world driving environments.

Practical deployment of language-in-the-loop driving must also address latency and compute constraints, since LLM inference can be too slow or expensive to run at the same frequency as a real-time planner. AsyncDriver proposes an asynchronous framework that decouples the LLM's inference rate from the planner's control loop, using language-derived guidance features while maintaining a high-frequency planning cycle. This line of work emphasizes that safe integration is not only about semantic correctness but also about systems design choices (update rates, fallbacks, and trust boundaries) that determine when and how language is allowed to influence motion planning.

**METHODS**
**Semantic Scene Assessment: Open-Vocabulary Hazard Screening**
For the first system, we implement a lightweight driving hazard detection system that uses a VLM to produce a one-dimensional hazard signal over time. We leverage foundational models such as CLIP [9], which compares the embeddings of an image and natural language text to output their image-text similarity score. Therefore by feeding such a model a front-facing camera frame from a moving vehicle and a prompt such as "hazard on the road," the system outputs a similarity value at each time step, enabling low-latency, real-time hazard screening.

**Data Collection** We evaluate our approach on the COOOL benchmark which is a collection of around 200 different videos containing unexpected road hazards (e.g., a deer crossing, a pedestrian falling into the roadway, etc). Each video was manually annotated to indicate at what frame interval the hazard becomes visible (e.g., pedestrian walking their dog on the sidewalk) and when it becomes a driving hazard (e.g., dog runs onto the driving road). These videos were then grouped into seven coarse hazard types: animal, pedestrian, airborne/falling object, road debris, low visibility, emergency scene, and construction zone to analyze how well broad categories perform. Because COOOL clips are hazard-dense, we additionally collection nominal driving footage across suburban, urban, and highway settings to balance the evaluation set to approximately 50/50 hazard and non-hazard frames ($\approx 34,000$ frames each).

**Prompting and Score Construction** Unlike other hazard detection pipelines that attempt to localize and classify a hazard object, our method uses broad prompts across the entire semantic scene. This design supports "catch-all" detection behavior for out-of-distribution hazards without requiring labeling for each hazard type. In addition to the seven category prompts, we also include a general "hazard" prompt as well that encompasses all seven defined categories and is intended to evaluate whether it can catch any unlabeled road hazards. Then to reduce any false positives, each hazard prompt is paired with a negative prompt (e.g., "normal driving scene"), and we use the margin between the two CLIP scores as the hazard confidence signal. A small set of prompt phrasings

With the hazard confidence signal, we threshold the margin in CLIP's scoring space. CLIP computes its similarity score from the dot product between L2-normalized image and text embeddings, followed by a learned temperature (logit scale). The logit scale amplifies separations between prompts, so the resulting margins are larger than the original cosine-similarity value. For each category, we convert the margin signal into a binary hazard prediction using a tuned threshold. Thresholds for the seven categories are tuned on their corresponding labeled subsets, while the general hazard threshold is tuned across all hazard videos.

**Evaluation and Optimal Threshold** Given the annotated frames of when the hazard is present, we then define our test metrics on the temporal intersection-over-union (tIoU) between predicted hazard segments and ground-truth hazard segments. Then to take into account false positives, we also report negative tIoU, which measures agreement with non-hazard intervals. Ideally, a hazard detection system should contain high positive tIoU (accurate hazard timing) while maintaining high negative tIoU (stable non-hazard behavior). To balance these objectives, we define a combined score:

$$Global\ tIoU\ =\ 1 - \frac{\sqrt{(1 - <positive\ tIoU>)^2 + (1 - <negative\ tIoU>)^2}}{\sqrt{2}} \qquad \text{Equation (1)}$$

We then select the optimal threshold by sweeping over the observed CLIP signals for each prompt set at a step size of 0.001 and choose the value that maximizes Global tIoU.

**Global Representation Learning: Vision–Language Embeddings for Trajectory**
In the second example system integrating vision and language into driving safety, we adapted the Motion Transformer (MTR) framework, originally developed for the Waymo Motion Dataset, to the Waymo End-to-End Driving Dataset for vision-based trajectory planning. This dataset provides multi-view 2D camera images, past ego-vehicle kinematic history (positions, velocities, and accelerations over a 5-second horizon), and a high-level driving intent.

We designed a scene context encoder that integrates visual and motion information to capture safety-relevant scene semantics. Past trajectory information is encoded using a transformer with self-attention over the ego vehicle's recent kinematic history. Visual context is extracted using a Vision Transformer (ViT) [14] applied to a stitched panorama formed from the front three camera views. These motion and visual features are combined, either through feature concatenation or cross-attention, to form scene context embeddings.

A transformer-based decoder then generates future trajectories conditioned on both the scene context and the intended maneuver. The scene context embeddings serve as keys and values, while an embedding of the driving intent is used as the query in a cross-attention module. To explore the role of high-level semantic representations, we additionally evaluated variants in which the intent query was augmented with scene-level embeddings from foundation models such as CLIP and DINOv2. These embeddings provide semantic and visual priors about the driving scene, allowing the planner to condition its outputs on both geometric context and higher-level scene semantics.

**Human–Vehicle Interaction: Natural Language as a Behavioral Constraint**
In the third system, we examine natural language as a high-level behavioral constraint that guides planning decisions. In this approach, natural language instructions are used to specify short-horizon (approximately 5–10 s) driving intent in a manner that is consistent with real-world, safety-relevant passenger-to-driver communication. This is done by combining the doScenes [6] dataset with OpenEMMA, a multimodal open-source motion planning framework based on Waymo's EMMA model.

doScenes is a language-augmented extension of nuScenes that retroactively annotates instructions using the "taxi test" heuristic: What instruction would you (as a passenger) give to the ego driver to cause the observed motion? We integrate doScenes into OpenEMMA by injecting the instruction into the model prompt as a passenger directive. We run OpenEMMA under two conditions for each scene: (1) no instruction input and (2) instruction-conditioned input using doScenes passenger directives. For nuScenes scenes that contain more than one doScenes passenger directive, each instruction is independently evaluated and metrics are computed at the instruction level. For trajectory alignment, we measure the Average Displacement Error (ADE), the standard geometric distance between predicted and ground-truth trajectories.

To isolate the effect of language conditioning, we treat the doScenes utterance as the single controlled change to the OpenEMMA pipeline. Concretely, we insert each doScenes instruction into the model's prompt as a passenger directive. After, the prompt instructs the model to follow the passenger's request by default, but to override it when compliance would be unsafe, in which case the model should choose the safest alternative and provide a brief justification.

**RESULTS**
**Semantic Scene Assessment: Open-Vocabulary Hazard Screening**
We first report the optimal thresholds obtained for the CLIP ViT-L/14 hazard signals. For each category prompt (and the general hazard prompt), we sweep thresholds over the observed score range and select the value that maximizes Global tIoU. This balances hazard coverage (positive tIoU) against false-alarms (negative tIoU). Table I summarizes the resulting optimal thresholds and their corresponding tIoU breakdowns.

*Table 1.*
*Optimal threshold score for best scoring Global tIoU for each category on CLIP model "ViT-L-14"*

| Category | Optimal Threshold | Global tIoU | Positive tIoU | Negative tIoU |
|---|---|---|---|---|
| Hazard | 0.315 | 0.572 | 0.589 | 0.556 |
| Pedestrian | 0.536 | 0.538 | 0.370 | 0.828 |
| Animal | 0.454 | 0.657 | 0.532 | 0.874 |
| Airborne/Falling | 2.214 | 0.473 | 0.256 | 0.951 |
| Low Visibility | 2.903 | 0.765 | 0.667 | 0.988 |
| Emergency Scene | 1.975 | 0.318 | 0.037 | 0.949 |
| Construction | 5.759 | 0.563 | 0.381 | 0.989 |
| Road Debris | -0.871 | 0.451 | 0.230 | 0.898 |

In our findings, "low visibility" and "animal" prompts perform the best with a Global tIoU of 0.765 and 0.657 respectively. Based on the positive tIoU score, low visibility prompts cover around 67% of all frames when a hazard is present and animal prompts cover around 53%. In contrast, categories such as "emergency scenes" and "road debris" score among the lowest with a Global tIoU of 0.318 and 0.451 respectively. Despite the range of positive tIoU scores from the seven broad categories, they all each maintain a high negative tIoU score suggesting the model has a better understanding of when there is no hazard in the image frame. Lastly, the general "hazard" prompt was established to evaluate if CLIP understood driving hazard using prompts such as "a driving hazard on the road" or "a hazard in the roadway ahead". Although it did achieve a global tIoU score of 0.572, it is susceptible to false alarms about half of the time given that the negative tIoU score is 0.556.

After selecting the threshold for each prompt set by maximizing Global tIoU, we evaluate how these category-level detectors can be composed into a single hazard screening system. We test three fusion strategies:
1. **No-hazard (categories only):** The seven prompt categories are evaluated independently on each frame using their category-specific thresholds. Here, the system flags a hazard if **any** category exceeds its threshold
2. **With-Hazard (categories + general hazard):** The system now includes seven category prompts plus the general "hazard" prompt. Here, the system flags a hazard if **any** category exceeds its threshold
3. **Dual-Hazard (hazard-gated agreement):** The system flags a hazard **only if** the "hazard" prompt exceeds its threshold **and** at least one additional category prompt also exceeds its threshold

*Table 2.*
*Performance evaluation on different hazard detection systems*

| System | Global tIoU | Positive tIoU | Negative tIoU | Video-TPR | Video-TNR |
|---|---|---|---|---|---|
| 1.) No-Hazard | 0.596 | 0.554 | 0.643 | 0.96 | 0.34 |
| 2.) With-Hazard | 0.571 | 0.605 | 0.539 | 0.99 | 0.34 |
| 3.) Dual-Hazard | 0.589 | 0.532 | 0.654 | 0.92 | 0.70 |

Table 2 compares hazard detection systems based on how category level prompts are combined. The No-Hazard system, which does not include general hazard prompts and solely relies on category specific broad prompts achieves the highest Global tIoU of 0.596. The other systems Global tIoU score are similar, in which the difference in the system changes on how well it alerts positive and negative hazards. The No-Hazard and With-Hazard system is a trade off between its positive tIoU score and negative tIoU score. However by introducing a Dual-Hazard signal where we need confirmation from the general "hazard" and at least 1 other broad category to signal a hazard, it helps reduce false alarms as it scores the highest negative tIoU score.

The difference between the systems becomes most apparent when shifting from frame-level to video-level analysis. We define Video-TPR (True Positive Rate) as the system's ability to detect a hazard at any point within a known hazard clip, while Video-TNR (True Negative Rate) measures the system's ability to correctly remain silent throughout a normal driving clip. While all systems issue an alert across all hazard videos, the Dual-Hazard approach improves reliability by avoiding false alerts in 70% of non-hazard driving clips. This is a larger improvement over the 34% achieved by the other two configurations with a slight decrease in Video-TPR accuracy.

Lastly, a notable observation made is that subtle changes to the prompts can have a significant influence on performance. Although an exhaustive test was not conducted, we tested different prompt pairs for each category. For the construction zone, we found the best prompt pair to be "road work ahead" with "a road with nothing unusual" which resulted in the 0.564 Global tIoU score. Yet when the prompt pair changed to "construction zone" with "a clear lane with no hazard" the Global tIoU had a 45.7% decrease in performance with a final score of 0.305.

**Global Representation Learning: Vision–Language Embeddings for Trajectory**

We evaluated the effect of incorporating vision-language representations into vision-based trajectory planning on the Waymo End-to-End Driving Dataset. Performance was measured using Average Displacement Error (ADE) and the Rater Feedback Score (RFS) [16], which reflects agreement with expert human-rated trajectories in safety-critical driving scenarios.

The base MTR-VP planning model, which conditions trajectory generation on camera imagery, past motion history, and high-level intent, served as the reference system. We compared this model against variants that augmented the planning query with scene-level embeddings from CLIP and DINOv2 [12].

As shown in Table III, incorporating CLIP and DINOv2 embeddings resulted in higher ADE at both 3-second and 5-second prediction horizons compared to the base MTR-VP model, indicating reduced trajectory accuracy. In addition, Tables IV and V show that the vision-language–augmented model consistently achieved lower Rater Feedback Scores across overall performance and multiple scenario categories, including construction zones, pedestrian interactions, and road debris. Taken together, these results demonstrate that, under the current formulation, directly conditioning the trajectory planner on global vision-language embeddings does not improve and in fact degrades safety-aligned planning performance relative to the base vision-based model.

*Table 3.*
*Average Displacement Error (ADE) for different prediction horizons across methods on the test split.*

| Model | 3-second trajectory ADE top-1 | 5-second trajectory ADE top-1 |
| --- | --- | --- |
| MTR-VP (base model) | 1.4232 | 3.3485 |
| MTR-VP with CLIP & DINOv2 embeddings | 2.0142 | 4.3379 |

*Table 4.*
*Rater Feedback Score (RFS) across methods on Waymo Test split for various categories.*

| Model | Overall | Construction | Intersection | Pedestrian | Cyclist |
| --- | --- | --- | --- | --- | --- |
| MTR-VP (base model) | 7.3433 | 7.7111 | 7.6864 | 7.5096 | 7.3138 |
| MTR-VP with CLIP & DINOv2 embeddings | 6.7862 | 6.9871 | 6.7769 | 7.1561 | 7.0224 |

*Table 5.*
*Rater Feedback Score (RFS) across methods on Waymo Test split for additional categories.*

| Model | Cut-In | Special Vehicles | Single Lane | Multi-Lane | Debris |
| --- | --- | --- | --- | --- | --- |
| MTR-VP (base model) | 7.4170 | 7.5705 | 7.7170 | 7.2205 | 7.3794 |
| MTR-VP w/ CLIP & DINOv2 embeddings | 6.9529 | 6.8868 | 7.5726 | 6.3238 | 6.9512 |

**Human-Vehicle Interaction: Language as a Behavioral Constraint**

We next evaluate the third use case: natural language-conditioned autonomous planning, in which passenger-style instructions are used to directly influence motion planning behavior. Across the full evaluation set using doScenes-injected prompts with OpenEMMA, instruction-conditioned planning altered model behavior relative to the vision-only baselines. In the baseline condition, the planner occasionally produced extreme failure cases, including trajectories that drifted outside drivable regions or advanced when the ego vehicle should have remained stationary. These failures resulted in very large ADE values that dominated the mean error. However, incorporating passenger instructions suppressed these catastrophic failures, yielding a large reduction in mean ADE relative to the vision-only baseline.

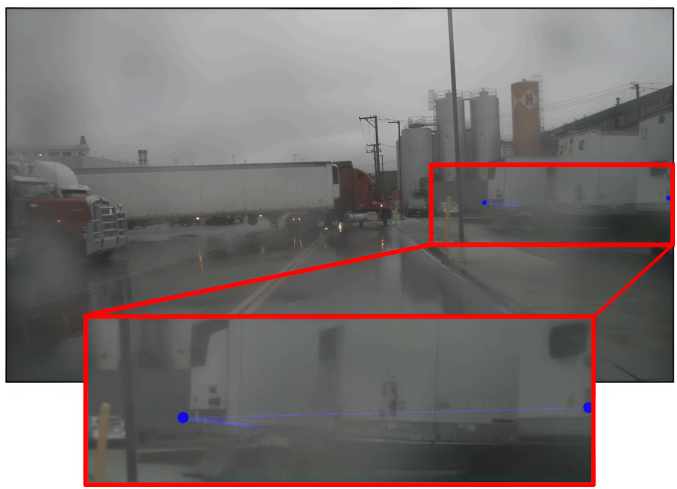

*Figure 1. Example of a high-error outlier in the visual-only baseline. The planner produces an out-of-bounds trajectory that departs from the drivable region, resulting in extremely large displacement error.*

As illustrated in Figure. 1, certain scenes resulted in planning failures in the visual-only baseline that can lead to trajectories that leave the drivable region. Although infrequent, these out-of-bounds behaviors incur extremely large displacement error and disproportionately affect mean ADE. We applied an upper 97.5th-percentile (Q97.5) filter to remove approximately twenty high-error outlier scenes. After filtering, instruction-conditioned planning achieved a more modest but consistent improvement. While average and poorly phrased instructions showed mixed effects, the best-performing instructions reduced mean ADE by 5.1% compared to the baseline. Additionally, we observe a slightly larger fraction of instructions outperform the visual-only baseline compared to the no-instruction condition.

*Table 6.*
*ADE comparison between visual only baseline condition and instruction conditions runs with all and outlier filtered scenes*

|  | Visual-only Baseline Avg ADE | With instruction Best ADE | With instruction Avg ADE | With instruction Worst ADE |
|---|---|---|---|---|
| Mean (All) | 6201.443 | 9.999 | 78.527 | 151.420 |
| Mean (Q97.5) | 2.879 | 2.732 | 2.929 | 3.11 |

As shown in Table 6, the large reduction in mean ADE across all scenes is primarily driven by the suppression of extreme outlier failures, while more modest improvements are observed after Q97.5 filtering. With the best-performing instruction resulting in the lowest average ADE of 2.732, we observe how different these trajectory predictions differ from a visual-only baseline. Figure 2 is a passenger instructions example that referenced observable scene elements and corrected unsafe baseline behavior. This example prevented the ego vehicle from proceeding through an active crosswalk as the instruction gave the model reasoning that the ego car should no longer proceed once it stops next to the curve. Multiple examples of corrections in unsafe behavior by the vehicle were still observed after the filtering of the outliers, which suggests that language can support in mitigating unsafe behaviors rather than uniformly improving trajectory accuracy.

**Scene-238**

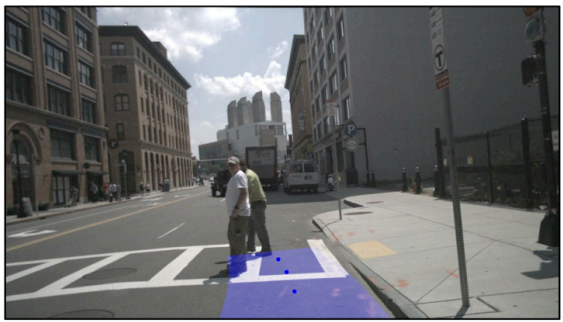
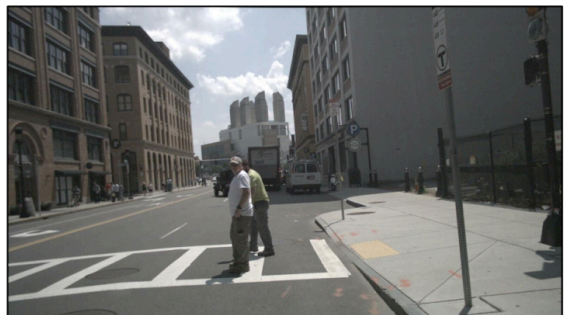

*No Instruction*        *"Stop at the curb on the right side of the road right before the crosswalk."*

*Figure 2. Qualitative comparison of trajectory predictions with and without natural language instruction.*

We further analyzed how instruction structure affected planning outcomes. We found that typical prompts of moderate length (approximately 9-12 words) yielded the largest improvements relative to the baseline, while shorter prompts underperformed. Furthermore, instructions that referenced dynamic scene elements, such as moving vehicles or pedestrians, achieved the lowest ADE on average. This suggests that prompt phrasing and grounding act as controllable variables that directly influence safety-aligned behavior.

**DISCUSSION**
The results from using a VLM (e.g. CLIP) as a general hazard screening model deployable in a real-time driving environment suggests certain strengths and weaknesses. Our experiments in Table 1 suggest that CLIP may have been trained more extensively on certain types of categories, such as "animal," resulting in a better understanding of animals within a driving scene. The model performed best in the category of low visibility, which included events such as dense fog or smoke covering the windshield where the recording camera was placed. This covers the majority of the scene and obstructs the view from the road which would match our prompt description (e.g. "low visibility on the road"). In contrast to low visibility performance, small hazards like "road debris" were among the lowest-scoring categories. This could be due to the encoding process, where the image is spatially downsampled, which may cause the model to lose smaller hazard details, leading it to classify the frame as a normal driving scene.

Another notable detail is that CLIP lacks temporal reasoning, which likely contributed to "emergency scene" being one of the lower-scoring categories. An emergency scene often involves flashing lights from first responders or hazard lights from a vehicle ahead. Because the standard off-the-shelf model only computes the similarity score for a single image-text pair, it evaluates frames in isolation. This makes it difficult to distinguish a vehicle with active flashing lights as it compares the prompt with either the lights of the vehicles being at an "off" state **or** "on" state. Addressing this limitation would require engineering the model to compare the text prompt across a sequence of frames rather than a single image.

The results from the Waymo End-to-End Driving experiments provide insight into the role of vision-language representations in safety-critical trajectory planning. While the base MTR-VP framework demonstrates that vision-based planning conditioned on past motion and intent can produce reasonable driving behaviors, the addition of global vision-language embeddings did not improve planning performance under the evaluated formulation. This suggests that the benefits of semantic representations learned by large vision-language models do not automatically transfer to low-level trajectory generation tasks without careful engineering of informative subfeatures from the general vision-language embeddings.

This lack of performance improvement from CLIP-based embeddings may also stem from a mismatch between the global, semantic nature of vision-language representations and the spatially grounded requirements of trajectory planning. While CLIP encodes high-level scene semantics, it does not explicitly preserve geometric relationships or agent-relative spatial structure, which are essential for generating safe and precise motion trajectories. When injected directly into the planning module, these global semantic cues may introduce ambiguity or noise rather than actionable guidance.

These findings highlight an important distinction between perception-level understanding and planning-level decision-making in autonomous driving systems. While vision-language models offer powerful tools for semantic reasoning and interpretation, their effective use in planning likely requires intermediate representations or structured grounding mechanisms that bridge high-level semantics with spatially grounded, geometry-aware decision processes.

In instruction-conditioned planning, we observed that different phrasings-expressed with the same intended actions-can lead to materially different trajectories. Instructions that clearly articulate a single action or reference observable scene element tend to produce better trajectory alignment, while vague or underspecified phrasing often degrades the performance. This highlights an underlooked aspect of instruction-conditioned planning that has direct implications for safety.

The doScenes-prompted OpenEMMA results highlight the distinction between using language as a global semantic representation (such as CLIP-style scene embeddings) and using it as an explicit, human-authored constraint on motion planning. In our experiments, passenger-style natural language instructions primarily acted as behavioral constraints, shaping vehicle responses in ambiguous scenes and suppressing rare but severe planning failures.

A key observation from the doScenes experiments found that instruction-conditioned planning supported with mitigating catastrophic failure cases. In several scenarios, the vision-only baseline produced unsafe behaviors, such as advancing the ego vehicle at a curbside crosswalk when it should have remained stationary. In some cases, scenes that involved active crosswalks or stopped lead vehicles, the baseline planner generated forward trajectories despite the presence of pedestrians or blocked right of way. In passenger instructions that explicitly referenced observable scene elements, such as the pedestrians, crosswalks, or traffic signals, this corrected these behaviors by constraining the planner to wait or stop. From a safety perspective, preventing such failures may be more consequential than marginal games in nominal trajectory accuracy.

At the same time, the suppression of catastrophic failures through language constraints introduce an important tradeoff: over-cautious behavior can also be unsafe or impractical, at times. A planner that overweights "stop" or "wait" language can become overly conservative, leading to hesitation, excessing yielding, or freezing in situations where proceeding is safe and expected. Avoiding collisions in isolation is not the only focus in driving strategy, but also about maintaining predictable, stable interaction with other road users. For example, overly conservative responses can increase rear-end risk through unnecessary braking, create deadlock at intersections, or produce socially noncompliant behavior that confuses surrounding drivers. In other words, language constraints that reduce one class of failures can introduce a different class of failures if they are applied without calibration. One practical approach is to interpret instructions as "soft constraints" that must still satisfy a traffic-rule logic, with explicit revalidation as the scene evolves.

In real-world passenger situations, it is expected that passengers do not always speak in clean or consistent language. Phrases may be short, words may be repeated, or instructions may be phrased in ways that are clear to a human, but ambiguous to a model (i.e., saying "right here" or "over there" instead of explicitly naming a lane, vehicle, or landmark). Spoken interfaces can further introduce noise through misheard words or transcription errors, which may shift the interpreted instruction away from the passenger's original intent without the passenger being aware of it. If a small change in wording can alter the trajectory of a motion vehicle, then it is possible that a passenger can unintentionally influence vehicle behavior even when their intent is harmless. For language-conditioned planning to be reliable in safety-critical settings, systems must focus on extracting the passenger's intended action and grounding that intent in the visible scene, rather than reacting to the surface form of the instruction alone.

**LIMITATIONS**

This study evaluates vision-language representations only through direct conditioning of a trajectory planning module, and does not explore alternative integration strategies such as object-level grounding, intermediate scene abstractions, or hierarchical planning architectures that may better leverage semantic information.

**Instruction Conditioned Planning**

While instruction-conditioned planning demonstrates the potential for natural language to act as a behavioral constraint, several limitations remain. The doScenes instructions are retrospectively authored using the "taxi test" heuristic rather than collected from real-time passenger interactions. Although this design enables controlled analysis, it introduces a distribution gap relative to spontaneous, spoken instructions that may be noisy, ambiguous, or incomplete. Real-world deployment would require robust mechanisms for intent extraction, ambiguity, resolution, and safety arbitration.

In addition, our evaluations were performed in an open-loop setting on logged nuScenes clips, where the ego vehicle behavior is already safe and successful. As a result, the planner is never required to recover from its own mistakes or operate under compounding error over time. Furthermore, all evaluated instructions are positive-goal-oriented, reflecting reasonable passenger intent. The system is not tested under unsafe, contradictory, or malicious instructions, nor under scenarios where language conflicts with traffic rules or physical feasibility.

**Trust Boundary**

We also do not define a production decision policy for how language inputs are accepted, rejected, or weighted relative to perception and planning constraints. Any real deployment would require a clear trust boundary, conservative fallbacks, and calibrated thresholds that determine when language can influence behavior and when it must be overridden.

**CONCLUSIONS**

Safety in open-world driving is not only a geometry problem. Many failures come from missing context: what is unusual, what is risky, and what a human would do in the same situation. Vision language models have recently emerged as a practical way to represent that context because they connect what the system sees with the words humans use to describe hazards and intent. This work tested a simple question with three system-level case studies: when do vision-language representations improve safety, and when do they fail to help?

In our first case study, we studied open-vocabulary hazard screening using CLIP-style image text similarity as a low latency semantic risk signal. The results show that prompt choice and structure matter. Category prompts reduce false alarms relative to a single broad 'hazard' prompt, and they work best when the hazard causes a clear scene-level change (for example, visibility hazards). However, framewise similarity remains brittle for hazards that are small, localized, or strongly temporal (for example, flashing emergency lights). The takeaway is that this approach is best positioned as a conservative "screening layer" that flags possible risk, not as a standalone detector. For real deployment, it must be engineered for calibration, composition, and temporal stability if it is to be dependable in safety-critical use.

Second, we tested whether global vision language embeddings improve planning when injected directly into a vision-based trajectory planner on the Waymo End-to-End Driving Dataset. In our evaluation, adding global CLIP or DINOv2 embeddings degraded both displacement error and human-aligned planning judgements. Ultimately, this negative result is important because it highlights a common trap: semantic representations are not automatically useful control features. A planning model needs grounded structure, such as explicit representations of where agents are, how they move, and which regions are drivable, not just more information. If semantics are to help planning, they likely need to be introduced in a more targeted form, for example through object-level grounding, intermediate scene abstractions, or hierarchical decision layers that can translate meaning into constraints and priorities.

Third, compelling safety benefits emerged when natural language was used as an human-authored constraint on motion planning. In doScenes-based instruction-conditioned planning with OpenEMMA, passenger-style language suppressed rare but severe planning failures and guided safer behavior in ambiguous situations. The value here was not that language made the planner "smarter" in general, but that language made behavior more legible and more conservative when the scene was underspecified, which is exactly where failures become costly.

Looking forward, the core challenge is not whether vision language models can describe a scene or, but whether we can consistently and reliably convert semantics into safe action. This points to several concrete directions: (i) temporally aware hazard screening that is stable across frames and calibrated for false alarms, (ii) structured semantic interfaces to planning what expose what the system believes is risky and why, instead of injecting undifferentiated embeddings, and (iii) safety arbitration for language in the loop, including robustness to unclear, conflicting, or unsafe instructions. More broadly, as autonomous vehicles move toward deployment in increasingly complex environments, the ability to reason about intent, ambiguity, and context (as or better than a human) will become as important as precise trajectory optimization. Vision-language representations are a strong foundation for this capability, but realizing safety benefits is an engineering problem that demands careful integration, grounding, and evaluation.